\documentclass[10pt,twocolumn,letterpaper]{article}

\usepackage[pagenumbers]{cvpr}

\usepackage{graphicx}
\usepackage{verbatim}
\usepackage{amsmath}
\usepackage{amssymb}
\usepackage{array}
\usepackage{multirow}
\usepackage{booktabs}
\usepackage{adjustbox}
\usepackage{arydshln}
\usepackage{enumitem}
\usepackage{textcomp}
\usepackage{pifont}
\usepackage{xspace}
\usepackage{xcolor}
\usepackage{algorithm}
\usepackage{algpseudocode}

\definecolor{oriGreen}{RGB}{0,220,117}
\definecolor{oriRed}{RGB}{220,0,0}
\definecolor{oriYellow}{RGB}{220,220,0}
\definecolor{oriCyan}{RGB}{0,220,220}
\definecolor{arrowcolor}{HTML}{BE4650}

\makeatletter
\DeclareRobustCommand\onedot{\futurelet\@let@token\@onedot}
\def\@onedot{\ifx\@let@token.\else.\null\fi\xspace}
\makeatother

\def\eg{\emph{e.g}\onedot}

\newcommand{\reffig}[1]{Fig.~\ref{#1}}

\newcommand{\refsec}[1]{Section~\ref{#1}}

\def\realBenchmarkName{OB-LIGM}
\def\sytheticBenchmarkName{OB-FUTURE}
\def\wacvsyntheticBenchmarkName{OB-Hypersim}
\def\finalRealBenchmarkName{\emph{RealOOB}}
\def\testMethodNum{forty}

\definecolor{cvprblue}{rgb}{0.21,0.49,0.74}
\usepackage[pagebackref,breaklinks,colorlinks,allcolors=cvprblue]{hyperref}

\title{RealOOB: A Definition-Consistent Real-World Oriented Occlusion Boundary Benchmark}

\author{
Lintao XU\textsuperscript{1} \quad
Yinghao WANG\textsuperscript{2} \quad
Chenchu RONG\textsuperscript{3} \quad
Xuchong QIU\textsuperscript{4}$^{\dagger}$ \quad
Chaohui WANG\textsuperscript{1}$^{\dagger}$
\\[0.6em]
{\small \textsuperscript{1} LIGM, UGE, ENPC, CNRS, France} \quad
{\small \textsuperscript{2} INFRES, Télécom Paris, IP Paris, France} \quad
{\small \textsuperscript{3} MathMagic, China} \quad
{\small \textsuperscript{4} Aptiv, China}
}

\begin{document}
\maketitle

\begin{abstract}
Occlusion boundaries (OBs) are pixel-level image boundaries corresponding to surface visibility discontinuities caused by occlusion. Through precise boundary localisation and occlusion orientation, OBs encode local surface layout and depth ordering, providing geometry-driven mid-level cues for scene understanding. However, progress in pixel-level OB estimation has been limited by \emph{fragmented supervision}: 
Existing benchmarks often suffer from limited coverage, category-specific designs, missing self-occlusion annotations, or inconsistent annotation definitions.
Meanwhile, modern edge detectors and monocular depth estimators have become strong boundary and geometry predictors, yet their relationship to definition-consistent OBs remains underexplored.
We introduce \finalRealBenchmarkName{}, a carefully annotated real-world benchmark with 4.26M definition-consistent, geometry-grounded OB labels covering both inter-object and self-occlusion boundaries, together with validity-aware occlusion-orientation maps that restrict supervision to pixels whose cross-boundary depth ordering is reliably measurable.
Based on \finalRealBenchmarkName{}, we evaluate \testMethodNum{} OB estimators and edge detectors alongside six monocular depth estimators.
Our evaluation reveals a clear gap in occlusion reasoning: modern edge detectors perform competitively with OB methods in localisation, whereas orientation prediction remains challenging for all evaluated methods. Meanwhile, even strong depth estimators often fail to exhibit measurable geometry at true OBs.
We believe \finalRealBenchmarkName{} provides a strong reference benchmark for the OB estimation community and a real-world testbed for assessing depth discontinuities and geometry fidelity in broader low-level vision tasks. Dataset and code will be released.
\end{abstract}

\section{Introduction}
\label{obb_sec:introduction}

Occlusion is a viewpoint-dependent phenomenon that arises when a 3D scene is projected onto a 2D image plane. Surfaces closer to the camera block farther surfaces along shared lines of sight, producing discontinuities in both the visible surface layout and the projected depth field. These events appear as either \textbf{\emph{inter-object occlusion}}, where one object occludes another, or \textbf{\emph{self-occlusion}}, where an object hides part of its own geometry. Occlusion is therefore both a source of ambiguity and a source of geometric information for scene understanding, with direct relevance to depth estimation~\cite{Zhang2022Self, Chen2023Self}, segmentation~\cite{Zhu2017Semantic, Qi2019Amodal}, object detection and tracking~\cite{Saleh2021Occlusion, Stadler2021Improving}, 3D reconstruction~\cite{karsch2013boundary, Popov2020CoReNet}, and human pose estimation~\cite{Sarandi2018how, Fu2015beyond, Cheng2019occaware}.

\begin{figure*}[t]
  \centering
  \vspace{-17pt}
  \includegraphics[width=0.9\linewidth]{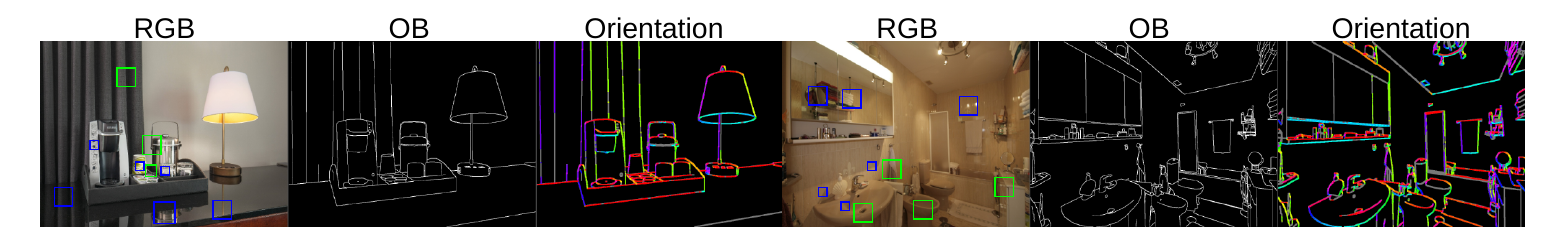}
  \vspace{-9pt}
    \caption[Examples from \finalRealBenchmarkName.]{
    \textbf{Examples from \finalRealBenchmarkName.}
    \textcolor{green}{Green} boxes highlight annotated OBs, including self-occlusion and fine-grained surface discontinuities; \textcolor{blue}{blue} boxes mark non-OB edges such as texture, illumination, reflection, normal-discontinuity edges without surface overlap, and objects seen through transparent surfaces.
    In the validity-aware orientation map, valid pixels are dilated for visibility, with hue encoding the depth-nearer side (\textcolor{oriGreen}{green}: left, \textcolor{oriRed}{red}: below, \textcolor{oriYellow}{yellow}: right, \textcolor{oriCyan}{cyan}: above); black/gray denotes non-OB or OB pixels without valid orientation.
    All OB annotations are semantic-free, following a geometry-grounded definition based solely on surface visibility discontinuity.
    }
    \vspace{-13pt}
  \label{fig:obb_teaser}
\end{figure*}

\textbf{Occlusion boundaries} (OBs, see~\reffig{fig:obb_teaser}) provide a compact image-domain representation of these visibility transitions: they are sparse binary boundary labels rather than dense surface-normal or depth maps. 
A binary OB map localises pixels at visible-surface discontinuities~\cite{wang2020occlusion}, while an oriented OB representation~\cite{wang2016doc,qiu2020P2ORM} additionally identifies which side of the boundary belongs to the nearer, occluding surface---the \emph{cross-boundary depth ordering}.
Unlike generic image edges, OBs are defined by scene geometry rather than appearance contrast: texture, shadow, reflection, and illumination edges are excluded unless they correspond to a true surface-visibility discontinuity. This makes OBs valuable geometry-driven mid-level cues for recovering scene structure and depth ordering.

Despite its long history in computer vision, OB estimation remains constrained by \textbf{\emph{fragmented supervision}} rather than by a simple lack of data. Existing real-world benchmarks vary substantially in annotation protocol: some are object-centric or category-dependent, many omit self-occlusion entirely or annotate it only sporadically, and depth-derived annotations primarily capture salient measurable depth discontinuities rather than the visibility-induced surface discontinuities. As illustrated in~\reffig{fig:previous_real_occ}, publicly available real-world OB annotations frequently miss self-occlusion and scene-structure boundaries, while including edges that do not satisfy a geometry-grounded OB definition. These inconsistencies make \emph{fair comparison} difficult and obscure whether improvements reflect better occlusion reasoning or adaptation to dataset-specific labelling rules.

To close this gap, we introduce \finalRealBenchmarkName{}, a carefully verified real-world benchmark with 4.26M exhaustive pixel-level OB annotations and validity-aware orientation labels under a unified, geometry-grounded definition~\cite{wang2020occlusion}. The benchmark covers both inter-object and self-occlusion boundaries, drawing from multiple complementary sources~\cite{vasiljevic2019diode,qi2022entity,wang2016doc,koch2018ibims1,honauer2016dataset} that span complex indoor scenes and texture-rich object-centric images. Its scale is comparable to widely used manually annotated edge and boundary benchmarks, while its annotation protocol is explicitly tailored to definition-consistent OB evaluation.

The methodological landscape around OB estimation has also shifted substantially. OB methods such as \cite{wang2016doc,feng2021mtorl} were developed on category-dependent datasets without self-occlusion annotations. Meanwhile, edge detection has advanced rapidly, spanning CNN-based detectors such as \cite{Xie2015Holistically,Liu2017Richer,he2019bdcn}, and recent ones using transformers or diffusion models~\cite{Pu2022EDTER,Ye2024DiffusionEdge,Cheng2026MEMO}. This raises a natural question: how far can modern edge detectors go as OB estimators, and where do they fail relative to dedicated OB methods? Our unified evaluation of \testMethodNum{} trained models shows that a repurposed edge detector with a lightweight orientation head achieves the best overall oriented-OB scores (B-ODS $0.764$, O-ODS $0.496$), while the same detector in boundary-only mode reaches even higher localisation (B-ODS $0.812$). A substantial gap between boundary and orientation scores persists across all methods, indicating that occlusion orientation estimation remains a major open challenge.

Beyond OB estimation, \finalRealBenchmarkName{} is also relevant to broader low-level dense prediction, particularly monocular depth estimation, where recent models increasingly emphasise boundary sharpness~\cite{bochkovskii2025depthpro,wang2025mogev2,xu2025ppd,yu2026infinidepth}. Since OBs correspond to surface-visibility discontinuities, a definition-consistent OB benchmark offers a natural testbed for probing a key open question: does visual sharpness translate to accurate boundary localisation, and how faithful is the predicted geometry near true OBs? Our experiments show that even on the depth-favourable evaluation subset, the best strict-matching F1 reaches only $0.260$, and the best joint cross-boundary depth-ordering accuracy is $74.4\%$, revealing that visually sharp depth maps do not yet yield pixel-accurate OBs or fully reliable cross-boundary geometry.

In summary, our contributions are threefold:

\begin{enumerate}

    \item We introduce \finalRealBenchmarkName{}, the \emph{first} benchmark with definition-consistent oriented occlusion labels on real-world-dominated images, exhaustively annotated and covering both inter-object and self-occlusion boundaries. It provides validity-aware orientation labels that restrict orientation supervision and evaluation to pixels with reliably measurable cross-boundary depth ordering.

    \item We train and evaluate \testMethodNum{} models---dedicated OB estimators and edge detectors---on \finalRealBenchmarkName{} under a unified protocol, showing edge detectors are strong OB baselines while oriented OB estimation remains a bottleneck.
    
    \item We use \finalRealBenchmarkName{} to probe 6 state-of-the-art depth estimators for boundary localisation and cross-boundary geometric fidelity, establishing it as a complementary real-world testbed for depth boundary evaluation.

\end{enumerate}
\section{Related Work}
\label{obb_sec:related_work}

\begin{figure*}[t]
    \centering
    \vspace{-13pt}
    \includegraphics[width=0.88\linewidth]{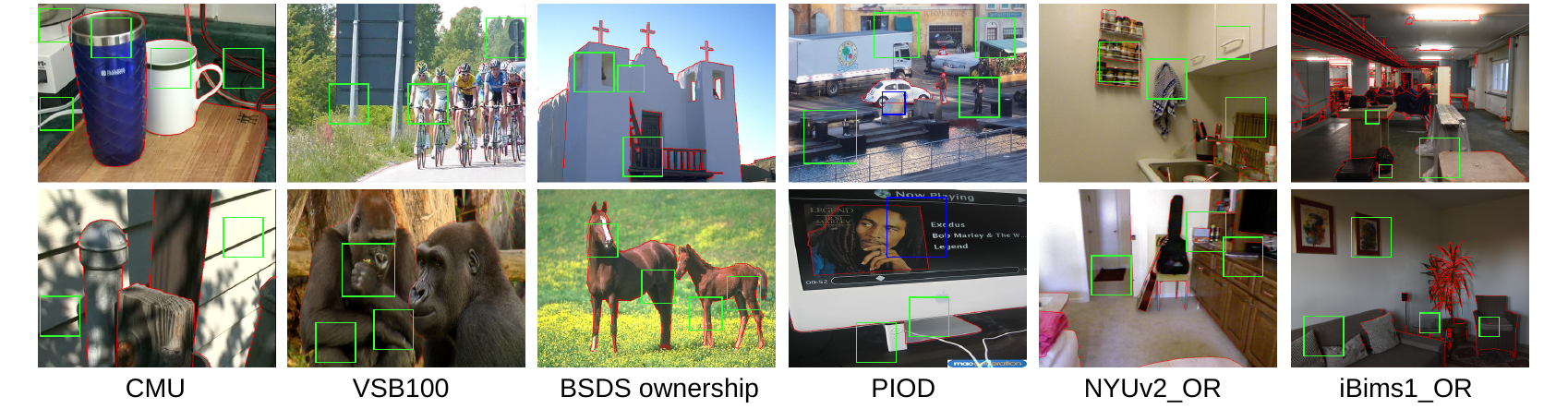}
    \vspace{-9pt}
    \caption[Previous real-world OB benchmark visualisation.]{
    \textbf{Publicly available real-world OB benchmarks}: CMU~\cite{stein2009occlusion}, VSB100~\cite{sundberg2011occlusion,galasso2013unified}, BSDS ownership~\cite{ren2006figure}, PIOD~\cite{hariharan2011semantic,wang2016doc}, NYUv2\_OR~\cite{ramamonjisoa2020predicting,qiu2020P2ORM}, and iBims1\_OR~\cite{qiu2020P2ORM}. \textcolor{red}{Red} overlays indicate annotated boundaries. \textcolor{green}{Green} boxes indicate missed OBs; \textcolor{blue}{blue} boxes indicate labelled edges that do not satisfy the geometry OB definition in~\cite{wang2020occlusion}. 
    }
    \vspace{-13pt}
    \label{fig:previous_real_occ}
\end{figure*}

\noindent\textbf{OB datasets.}
Existing OB datasets differ in annotation protocol, coverage, and underlying definition. Early real-world benchmarks~\cite{stein2009occlusion,sundberg2011occlusion,ren2006figure,hoiem2011recovering} are limited in scale, lack orientation labels, or only partially cover self-occlusion. PIOD~\cite{wang2016doc} provides larger-scale oriented annotations but is restricted to 20 object categories, so OBs outside the 20 categories, including many scene-structure and fine-grained self-OBs, are not exhaustively annotated. Depth-derived OB annotations such as~\cite{qiu2020P2ORM} inherit noise and incompleteness from the source depth, capturing salient depth discontinuities. Synthetic datasets are complementary: \sytheticBenchmarkName~\cite{xu2024iobe} offers accurate self-occlusion-aware OBs in simple indoor scenes, while \wacvsyntheticBenchmarkName~\cite{xu2026modot} adds photorealism with less complete self-occlusion coverage.
As shown in Fig.~\ref{fig:previous_real_occ}, real-world annotations consistent with a single geometric definition remain limited---a gap that \finalRealBenchmarkName{} addresses with exhaustive, self-occlusion-aware OB annotations and validity-aware orientation labels.

\noindent\textbf{OB estimation.}
Early OB methods relied on hand-crafted cues: video-based approaches combined motion discontinuities, temporal consistency, T-junctions, and graphical models~\cite{stein2006local,stein2009occlusion,apostoloff2005learning,feldman2007motion}; monocular approaches added geometric context and pseudo-depth~\cite{hoiem2007recovering,he2010occlusion}.
Deep learning shifted the field toward learned boundary-and-orientation estimation. DOC~\cite{wang2016doc} introduced the joint boundary-orientation formulation, followed by CNN refinements addressing class imbalance, multi-scale fusion, and multi-task supervision~\cite{wang2019doobnet,lu2019ofnet,feng2021mtorl,lu2019context,li2023fsinet,hambarde2024occlusion}. Geometry-aware variants include~\cite{qiu2020P2ORM,xu2024iobe} and joint depth--OB estimation~\cite{xu2026modot}.
Despite this progress, evaluation has largely relied on category-dependent datasets with limited self-occlusion coverage or synthetic supervision, motivating a definition-consistent real-world benchmark.

\noindent\textbf{Edge detection.}
OBs are generally regarded as a subset of image edges~\cite{stein2008thesis}, but edge detectors also respond to texture and illumination, whereas OBs are restricted to visibility-induced surface discontinuities~\cite{wang2020occlusion}---making OB localisation an open empirical test for edge detectors.
Deep edge detection has evolved from CNN-based detectors~\cite{Xie2015Holistically,Liu2017Richer,yu2017casenet,he2019bdcn,Su2021Pixel,soria2023dexined_ext,soria2022ldc} to methods with stronger global context, generative formulations, or ranking-based objectives~\cite{Pu2022EDTER,Jie2024EdgeNAT,Cetinkaya2024RankED,Li2025Doubly,Ye2024DiffusionEdge}.
A parallel line models multi-annotator uncertainty or multi-granularity boundaries~\cite{Zhou2023The,zhou2024muge,Liufu2025SAUGE}, while crisp edge detectors focus on sharper localisation with reduced NMS reliance~\cite{huan2021unmixing,Cetinkaya2026MatchED,Cheng2026MEMO}.

\noindent\textbf{Depth estimation and boundary sharpness.}
Earlier work refined depth near OBs with twin-surface representations~\cite{imran2021depth}, displacement fields~\cite{ramamonjisoa2020predicting}, or occlusion-aware modules~\cite{qiu2020P2ORM,hambarde2024occlusion}. 
Modern monocular depth estimation has advanced with large-scale pre-training and foundation models such as Depth Anything v1/v2~\cite{yang2024depthanything,yang2024depthanythingv2}, and recent methods further emphasise boundary sharpness~\cite{bochkovskii2025depthpro,wang2025mogev2,xu2025ppd,yu2026infinidepth}.
However, sharp depth edges do not necessarily imply accurate boundary localisation or faithful geometry around OBs: depth-derived edges can be misaligned or biased toward strong depth contrasts, while subtle occlusions with weak measurable depth gaps may be missed. 
\finalRealBenchmarkName{} provides a complementary testbed for evaluating depth boundary localisation and cross-boundary geometric fidelity.

\section{The \finalRealBenchmarkName{} Benchmark}
\label{obb_sec:datasets}

\begin{figure*}[t]
  \vspace{-13pt}
  \centering
  \begin{minipage}{\linewidth}
    \centering
    \includegraphics[trim=0 302bp 0 0, clip, width=0.8\linewidth]{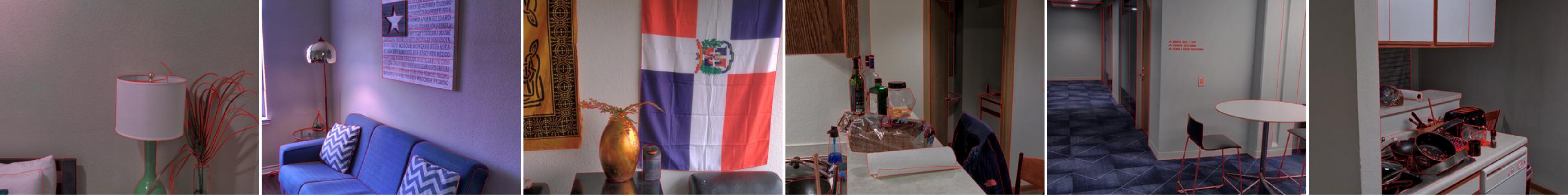}
  \end{minipage}
  \begin{minipage}{\linewidth}
    \centering
    \includegraphics[trim=0 0 0 302bp, clip, width=0.8\linewidth]{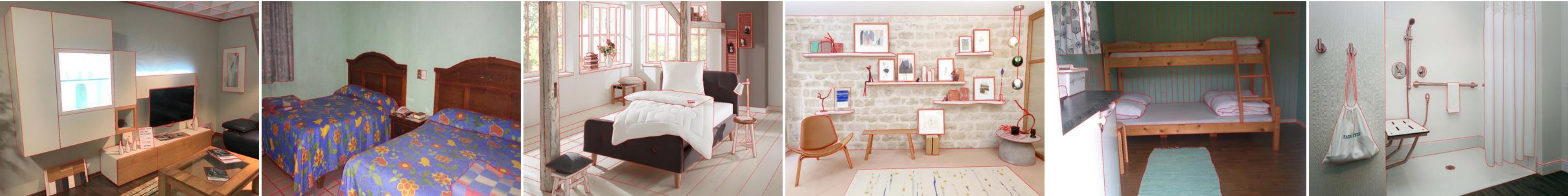}
  \end{minipage}
  \begin{minipage}{\linewidth}
    \centering
    \includegraphics[width=0.8\linewidth]{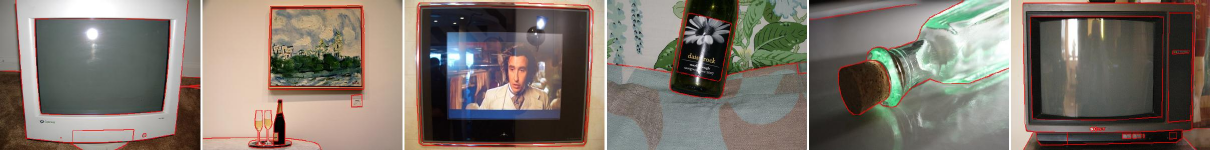}
  \end{minipage}
  \vspace{-6pt}
  \caption[Additional visualisations from \finalRealBenchmarkName.]{
  \textbf{More visualisations from \finalRealBenchmarkName.}
  \textcolor{red}{Red} overlays indicate annotated OBs, excluding intensity-based edges such as texture.}
  \vspace{-13pt}
  \label{obb_fig:vis_more}
\end{figure*}

\vspace{-2pt}
\subsection{Source Composition}
\label{obb_sec:sources}

\finalRealBenchmarkName{} contains 520 fully annotated images collected from five sources: DIODE~\cite{vasiljevic2019diode}, EntitySeg~\cite{qi2022entity}, PIOD~\cite{wang2016doc}, iBims-1~\cite{koch2018ibims1}, and LF4D~\cite{honauer2016dataset}. The benchmark subsumes and re-verifies the 120 images of \realBenchmarkName~\cite{xu2024iobe}. The benchmark is real-world-dominated with an indoor and object-centric focus (see~\reffig{obb_fig:vis_more}): complex indoor scenes with dense self-occlusion, and structurally simpler but texture-rich object-centric images whose abundant non-OB edges test false-positive (FP) suppression. A small number of synthetic or non-photographic samples are included for geometric and visual diversity. 

The sources provide complementary properties. DIODE contributes 200 high-resolution indoor images ($1024 \times 768$) with ground-truth (GT) depth, mainly annotated from scratch. EntitySeg contributes 271 images with diverse visual styles, resolutions, and object layouts (average resolution $952 \times 880$). PIOD contributes 40 object-centric images (average $439 \times 434$), while LF4D and iBims-1 contribute 9 geometry-rich examples with high-quality disparity or depth.
For all sources except DIODE, available segmentation, OB, depth-, or disparity-derived contours serve only as candidate boundaries. Final OB annotations are manually verified under the same geometry-grounded definition in~\cite{wang2020occlusion}: non-OB and inaccurate contours are removed, and missing object and self-occlusion boundaries are added.

In addition to the benchmark images, we release auxiliary resources, excluded from both the benchmark split and the reported statistics unless stated otherwise: 248 DIODE hard negatives---textured walls and ceilings with no OBs---used during training for FP suppression and segmentation-derived boundary assets with validity-aware orientations from 5{,}090 carefully selected EntitySeg images.

\vspace{-2pt}
\subsection{Annotation Protocol}
\label{obb_sec:annotation_protocol}

\begin{figure}[ht]
  \vspace{-6pt}
  \centering
  \includegraphics[width=0.88\linewidth]{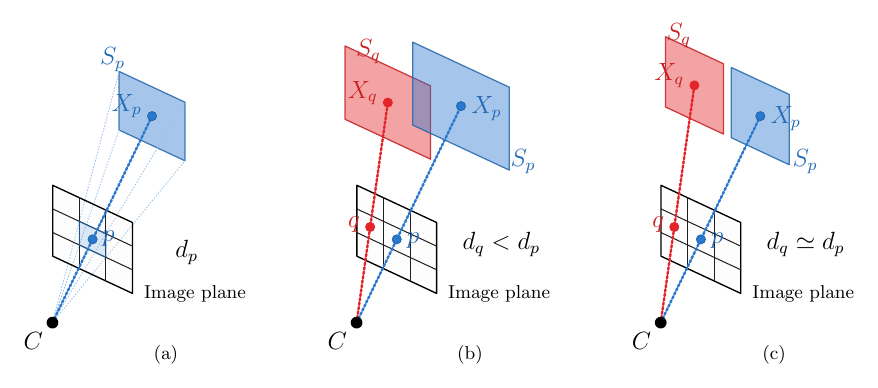}
  \vspace{-9pt}
  \caption[Geometric interpretation of pixel-level OB annotation.]{
  \textbf{Geometric interpretation of semantic-free pixel-level OB annotation.}
  (a) A pixel on the image plane represents the projection of a visible 3D surface patch along a camera ray $C$.
  (b) For two neighbouring pixels $p$ and $q$, their visible patches $S_p$ and $S_q$ define a local occlusion event when one surface occludes the other along the viewing direction; these patches may belong to different objects (inter-object occlusion) or to different visible parts of the same object (self-occlusion).
  We annotate a thin, single-pixel OB at the interface rather than labelling both neighbouring pixels.
(c) Where $S_p$ and $S_q$ lie on different surfaces that each occlude the background behind them but not each other, and whose depth step is too small to resolve at image resolution, the boundary is still a surface-visibility discontinuity, while its cross-boundary depth ordering is masked out from orientation supervision and evaluation.
  }
  \vspace{-6pt}
  \label{obb_fig:vis_ob}
\end{figure}

Annotations follow the formal, geometry-grounded OB definition in~\cite{wang2020occlusion}. It treats an OB as a collection of local occlusion events: a small boundary segment corresponds to an occlusion relation between the 3D surfaces projecting to the two neighbouring 2D image regions, and the full OB map is obtained by considering all such events over the image. The guiding principle for annotation is therefore \textbf{\emph{surface visibility discontinuity}} (see~\reffig{obb_fig:vis_ob}): an annotated boundary should correspond to a change in the visible surface caused by occlusion. Boundaries caused only by texture, illumination, reflection, or normal changes without surface overlap are excluded. Edges of objects visible through transparent media are treated as lying behind the transparent surface and are not annotated as OBs unless they correspond to an opaque surface-visibility discontinuity. Following~\cite{wang2020occlusion}, image-border pixels are not considered OBs; we exclude all annotations within a 2-pixel border and apply this convention consistently in statistics and experiments.

Ground-truth OBs are annotated using \emph{ByLabel}~\cite{Qin18}, a pixel-accurate edge-level annotation tool that supports open curves and fine boundary structures. Edge-level annotation is important here because self-occlusion boundaries may be non-closed and are poorly represented by polygon-based segmentation tools. \emph{Eight} annotators participated in the labelling process, and all annotations underwent multiple rounds of cross-checking. Annotation difficulty varies substantially: complex indoor scenes with dense self-occlusion may require more than five hours, whereas simple object-centric scenes can be annotated more quickly. Annotation and quality control averaged more than 1.5 hours per image.

\subsection{Validity-Aware Orientation Labels}
\label{obb_sec:gen_ori}

In addition to binary OB annotations, \finalRealBenchmarkName{} provides occlusion-orientation labels following the left-hand-rule convention used in prior oriented OB benchmarks~\cite{wang2016doc,ren2006figure}. Assigning dense orientation labels manually is substantially harder than drawing OB locations, particularly for tiny, subtle, or non-closed self-occlusion boundaries. 
Moreover, not every OB pixel admits an equally reliable cross-boundary depth-ordering relation: object contours and subtle self-occlusion boundaries are genuine surface-visibility discontinuities, but the depth step across them is often too small or too noisy to resolve at image resolution (see~\reffig{obb_fig:vis_ob} (c)).

\begin{figure}[ht]
  \centering
  \includegraphics[width=0.93\linewidth]{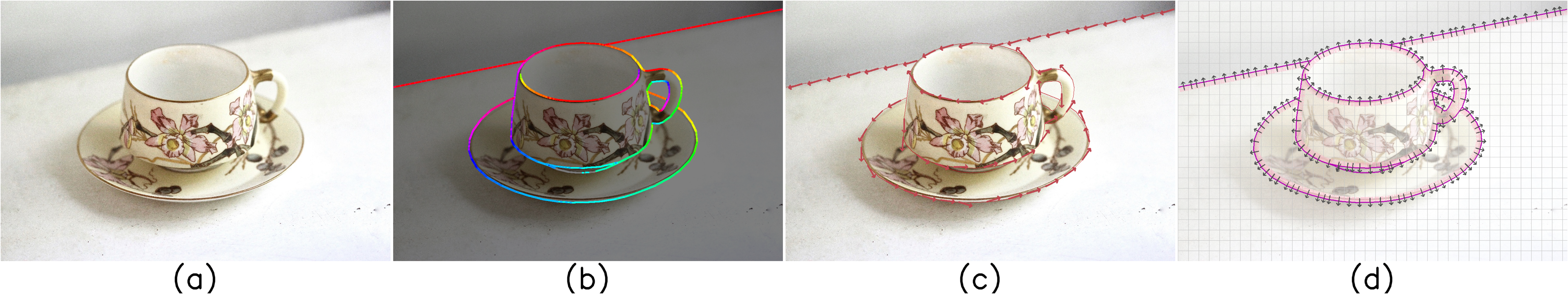}
  \vspace{-6pt}
  \caption[Visualisation of occlusion orientation labels in \finalRealBenchmarkName.]{
    \textbf{Visualisation of occlusion orientation.}
    (a) RGB image.
    (b) Orientation map with valid orientation pixels dilated for visibility; hue encodes the depth-nearer side as in \reffig{fig:obb_teaser}.
    (c) OB annotations with \textcolor{arrowcolor}{red arrows} showing orientation $\theta$ under the left-hand rule~\cite{wang2016doc}, where the depth-nearer side lies to the left of each arrow.
    (d) Local occlusion-ordering grid, where black arrows point from the depth-nearer to depth-farther surface across each boundary.
    Zoom in for better visibility.
    }
  \vspace{-6pt}
  \label{obb_fig:vis_ori}
\end{figure}

\begin{table*}[t]
\vspace{-13pt}
\centering
\caption[Summary of representative OB and edge/boundary datasets.]
{\textbf{Summary of representative OB and edge/boundary datasets with publicly available annotations.}
Edge/boundary datasets are included as contextual references but do not provide occlusion-specific annotations.
$^\dagger$: multiple annotations per image.  
$^\ddagger$: only for evaluation.
\emph{Orient.}, \emph{Self-Occ}, \emph{Exhaust.}, and \emph{Def.} denote occlusion orientation, self-occlusion coverage, exhaustive full-image annotation, and use of a formal geometry-grounded OB definition, respectively.}
\vspace{-6pt}
\label{table:obb_dataset_compare}
\resizebox{0.8\textwidth}{!}{%
\begin{tabular}{lllccccccl}
\toprule
\textbf{Dataset} & \textbf{Task} & \textbf{Scale} & \textbf{Avg. Resolution} 
& \textbf{Orient.} & \textbf{Self-Occ} & \textbf{Exhaust.} 
& \textbf{Def.} & \textbf{Domain} & \textbf{Aux Info} \\
\midrule
NYUD~\cite{silberman2012nyudv2} & Contour & 1{,}449 & $560 \times 425$ & N/A & N/A & \checkmark & N/A & Indoor & Geometry \\
Multicue$^\dagger$~\cite{mely2016multicue} & Contour/Edge & 100 & $1280 \times 720$ & N/A & N/A & \checkmark & N/A & Both & --- \\
BIPED~\cite{soria2023dexined_ext} & Edge & 250 & $1280 \times 720$ & N/A & N/A & \checkmark & N/A & Outdoor & --- \\
BSDS500$^\dagger$~\cite{arbelaez2010contour} & Edge & 500 & $432 \times 370$ & N/A & N/A & \checkmark & N/A & Outdoor & --- \\
BRIND~\cite{Pu_2021ICCV_brind} & (Geom.) edge & 500 &  $432 \times 370$ & N/A & N/A & \checkmark & N/A & Outdoor & --- \\
UDED$^\ddagger$~\cite{Soria2023Tiny} & Edge & 50 &  $531 \times 463$  & N/A & N/A & \checkmark & N/A & Both & --- \\
\midrule
CMU$^\ddagger$~\cite{stein2009occlusion} & OB & 30 & $546 \times 401$ & \texttimes & Partial & \texttimes & \texttimes & Both & Video \\
VSB100$^\dagger$$^\ddagger$~\cite{sundberg2011occlusion,galasso2013unified} & OB & 644 & $1358 \times 774$ & \texttimes & \texttimes & \checkmark & \texttimes & Both & Video \\
BSDS ownership$^\dagger$~\cite{ren2006figure} & OB & 200 & $321 \times 481$ & \checkmark & \texttimes & \checkmark & \texttimes & Outdoor & --- \\
PIOD~\cite{hariharan2011semantic,wang2016doc} & OB & 10{,}094 & $475 \times 390$ & \checkmark & \texttimes & \texttimes & \texttimes & Both & --- \\
NYUv2\_OR$^\ddagger$~\cite{ramamonjisoa2020predicting,qiu2020P2ORM} & OB & 654 & $592 \times 440$ & \checkmark & Partial & \texttimes & \texttimes & Indoor & Geometry \\
iBims1\_OR$^\ddagger$~\cite{qiu2020P2ORM} & OB &  100  & $640 \times 480$   & \checkmark &  Partial  & \texttimes  & \texttimes  & Indoor & Geometry \\
OB-LIGM$^\ddagger$~\cite{xu2024iobe} & OB &  120  & $998 \times 837$   & \texttimes &  \checkmark  & \checkmark  & \checkmark  & Indoor & Geometry \\
\midrule
\textbf{Ours \finalRealBenchmarkName{}} & OB & 520 &  \textbf{$933 \times 796$}
& \textbf{\checkmark} & \textbf{\checkmark} & \textbf{\checkmark} & \textbf{\checkmark} & Indoor & Geometry \\
\bottomrule
\end{tabular}%
}
\vspace{-13pt}
\end{table*}

We therefore provide \textbf{\emph{validity-aware}} orientation labels. For each annotated OB pixel, the orientation indicates the depth-nearer side only when the local depth ordering is reliably measurable from the available geometry; otherwise, the pixel remains part of the binary OB ground truth but is masked out for orientation supervision and evaluation. Thus, all annotated OB pixels are used for boundary localisation, while orientation losses and metrics are computed only on valid-orientation pixels.

Orientation labels are derived from available geometric cues: for images with GT depth, from local boundary normals and two-sided depth samples; for images without GT depth, from an agreement test over two independent monocular depth estimators~\cite{wang2025mogev2,xu2025ppd}. Predicted depth serves \textbf{\emph{only}} as evidence for the orientation at existing OB pixels, not for defining boundary locations. Pixels without sufficient depth evidence are masked out rather than forced into orientation supervision, and we further manually verify cases where the cross-boundary depth ordering is visually unambiguous. \reffig{obb_fig:vis_ori} (b) illustrates the orientation convention used.

Across \finalRealBenchmarkName{}, 2.72M of 4.26M OB pixels receive valid orientation labels, an overall valid rate of 63.9\%; per-source rates appear in the supplementary material (\emph{Supp.}).
These rates vary with data source, scene geometry, depth quality, and visual ambiguity: subsets with clean geometric evidence or visually unambiguous object-centric layouts tend to have higher valid rates, whereas complex indoor scenes may contain small, ambiguous, or noisy cross-boundary depth differences even when RGB-D data are available. Invalid-orientation pixels are \emph{not} discarded from the benchmark---they remain part of the binary OB GT and contribute to boundary localisation evaluation. Detailed orientation generation, parameter settings, and validation on the synthetic dataset~\cite{xu2026modot} are provided in \emph{Supp}.

\subsection{Dataset Analysis and Comparison}
\label{obb_sec:dataset_stats}

\noindent\textbf{Annotation properties.}
Table~\ref{table:obb_dataset_compare} compares \finalRealBenchmarkName{} with representative real-world OB datasets and common edge/boundary benchmarks. Existing OB datasets vary in annotation protocol: some are object-centric or category-dependent, some are derived from depth or semi-automatic procedures, and most lack exhaustive full-image annotations with self-occlusion coverage. Edge and contour datasets are included for reference but do not distinguish OBs from other edge types. Among these, \finalRealBenchmarkName{} is the only entry providing oriented, exhaustive, definition-consistent, and self-occlusion-aware OB annotations.

\begin{figure}[ht]
  \centering
  \includegraphics[width=0.88\linewidth]{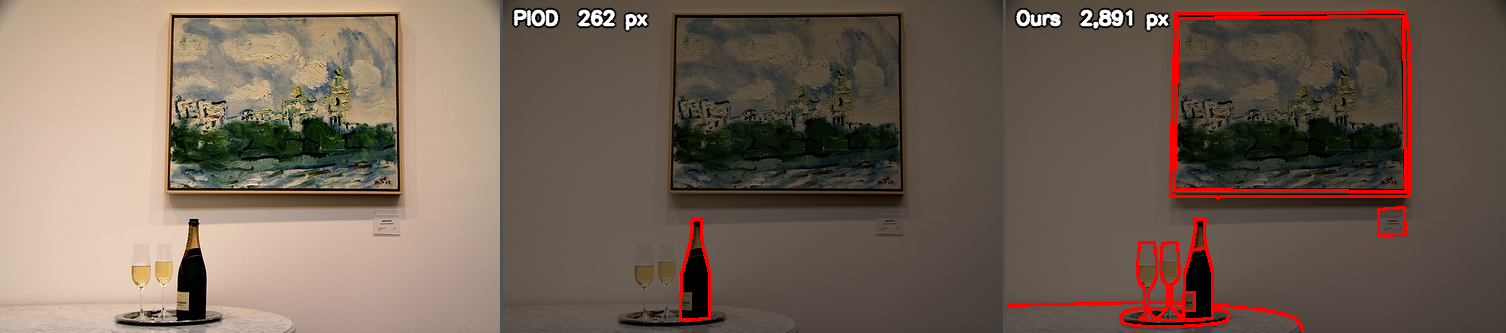}
  \vspace{-6pt}
  \caption[PIOD and \finalRealBenchmarkName{} annotations on the same image.]{
\textbf{PIOD vs.\ \finalRealBenchmarkName{} annotations on the same image.} Top-left: OB pixel counts. PIOD labels only the bottle, one of its 20 predefined categories; the painting, glasses, and table are left unannotated, and self-occlusion is excluded by design---despite its large scale, PIOD is not proper for semantic-free OB evaluation.}
  \label{fig:piod_gt_compare}
  \vspace{-13pt}
\end{figure}

\vspace{2pt}
\noindent\textbf{Per-image supervision.} 
Beyond image count, the amount of OB supervision depends on image resolution, annotation density, and whether the annotation is exhaustive and covers self-occlusion. Table~\ref{tab:dataset_comparison} compares \finalRealBenchmarkName{} with representative OB datasets under the same 2-pixel border-exclusion convention; for multi-annotator datasets, we report the per-annotation average. \finalRealBenchmarkName{} contains 4.26M OB pixels at 8.20K pixels per image on average, higher than PIOD, BSDS ownership, CMU, NYUv2\_OR, and iBims1\_OR, and VSB100.
The per-image OB count is lower than that of its OB-LIGM subset (8.20K vs.\ 10.5K) because \finalRealBenchmarkName{} includes structurally simpler object-centric images with fewer OBs but richer non-OB edges, which are important for evaluating FP suppression.
Although PIOD contains more images, its annotation protocol---which annotates only 20 predefined categories, is therefore non-exhaustive, and ignores self-occlusion---results in substantially fewer OB pixels per image (see~\reffig{fig:piod_gt_compare}); even on the 40 structurally simple PIOD images we re-annotated, its labels miss $46.7\%$ of our OB pixels.
Image count alone does not reflect the amount of definition-consistent OB labels that a dataset provides.

\begin{table}[h]
  \vspace{-3pt}
  \centering
  \caption[Quantitative comparison of OB datasets.]{
  \textbf{Quantitative comparison of OB datasets.} \emph{Density} and \emph{OB/Canny}: dataset-wide ratios of OB pixels to image pixels and to Canny edge pixels.}
  \vspace{-6pt}
  \label{tab:dataset_comparison}
  \resizebox{\columnwidth}{!}{
  \begin{tabular}{lrrrrr}
    \toprule
    \textbf{Dataset}  & \textbf{Scale}  & \textbf{Total OB} & \textbf{OB/Image} & \textbf{Density} & \textbf{OB/Canny} \\
    \midrule
    PIOD~\cite{hariharan2011semantic,wang2016doc}                          & 10{,}094  & 17.7M & 1.75K & 0.98\% &  5.0\% \\
    BSDS ownership~\cite{ren2006figure}                &      200  & 637K  & 3.19K & 2.11\% &  9.8\% \\
    CMU~\cite{stein2009occlusion}                           &       30  & 120K  & 4.00K & 1.74\% & 10.8\% \\
    NYUv2\_OR~\cite{ramamonjisoa2020predicting,qiu2020P2ORM}                     &      654  & 2.45M  & 3.74K & 1.46\% & 15.1\% \\
    iBims1\_OR~\cite{qiu2020P2ORM}                    &      100  & 635K  & 6.35K & 2.10\% & 22.8\% \\
    VSB100~\cite{sundberg2011occlusion,galasso2013unified}                        &      644  & 5.16M  & 8.01K & 0.75\% &  7.7\% \\
    OB-LIGM~\cite{xu2024iobe}                        &      120  & 	1.25M  & 10.5K & 1.28\% &  14.7\% \\
    \midrule
    \textbf{Ours \finalRealBenchmarkName{}}                 & 520  & 4.26M & 8.20K & 1.10\% & 12.7\% \\
    \bottomrule
  \end{tabular}%
  }
  \vspace{-6pt}
\end{table}

\noindent\textbf{OBs vs.\ edges.}
To quantify the relation between OBs and image edges, we compare \finalRealBenchmarkName{} with Canny edge maps~\cite{canny}. With a 2-pixel tolerance, \textbf{91.7\%} of OB pixels lie near Canny edges, supporting the view that OBs are largely a subset of image edges~\cite{stein2008thesis}. Conversely, the total number of OB pixels is only \textbf{12.7\%} of the number of Canny edge pixels, indicating that intensity-based edge maps contain many appearance-induced edges unrelated to occlusion. This gap motivates evaluating edge detectors on \finalRealBenchmarkName{}: they provide strong boundary-localisation baselines but must also suppress geometrically irrelevant edges to perform well on OB estimation.

\vspace{2pt}
\noindent\textbf{Train/Test Split.}
We partition \finalRealBenchmarkName{} into 390 training and 130 test images using a stratified 75/25 split by data source for OB benchmarking. 
Monocular depth estimators involve no training split, and are evaluated on all 520 images or on a GT-depth subset. 
The two partitions have closely matched orientation-angle distributions; per-source allocations and detailed split statistics appear in \emph{Supp.}

\section{Benchmark Experiments}
\label{obb_sec:experiments}

We benchmark dedicated OB estimation methods, modern edge detectors, and monocular depth estimators on \finalRealBenchmarkName{}. The experiments are designed to answer three questions: (i) how existing OB methods perform under a unified real-world protocol and (ii) whether modern edge detectors can serve as strong OB localisation baselines (\refsec{obb_sec:results}); and (iii) whether sharp monocular depth predictions recover definition-consistent OBs (\refsec{sec:depth-edge-protocols}).

Due to space constraints, more experimental setup, ablations and qualitative results, cross-dataset experiments across \finalRealBenchmarkName{}, synthetic and prior OB datasets, 
and an analysis of what prior OB benchmarks measure and how their rankings differ from ours are provided in \emph{Supp.}

\subsection{Experimental Setup for OB Estimation}
\label{obb_sec:experimentalSetup}

\noindent\textbf{Baselines.}
We evaluate three groups of methods. First, \textbf{\emph{dedicated OB methods}} jointly predict OB and orientation, including~\cite{wang2016doc,wang2019doobnet,lu2019ofnet,qiu2020P2ORM,lu2019context,feng2021mtorl,xu2024iobe,li2023fsinet,hambarde2024occlusion,xu2026modot}.
Second, we train modern \textbf{\emph{edge detectors}} in their original boundary-only form, covering CNN-, transformer-, diffusion-, and crisp-edge models~\cite{Xie2015Holistically,Liu2017Richer,yu2017casenet,he2019bdcn,poma2020biped,Su2021Pixel,huan2021unmixing,soria2022ldc,Pu2022EDTER,Cetinkaya2024RankED,Li2025Doubly,li2026nbed,Ye2024DiffusionEdge,Cheng2026MEMO,Cetinkaya2026MatchED}. 
Third, we adapt \textbf{\emph{edge detectors with a lightweight orientation head}} inspired by OPNet~\cite{feng2021mtorl}: the boundary branch keeps its native architecture and edge loss, while the added head predicts occlusion orientation from multi-level features and is supervised with the OPNet orientation loss.
For dedicated OB methods and boundary-only edge detectors, we keep the original architectures and losses whenever possible, changing only the data and training protocol.

\vspace{2pt}
\noindent\textbf{Metrics.}
We follow the standard precision--recall (PR) protocol (SEval) used in OB and edge detection~\cite{wang2016doc,feng2021mtorl,he2019bdcn,Pu2022EDTER}, reporting B-ODS, B-OIS, B-AP for boundaries and O-ODS, O-OIS, O-AP for orientation; orientation metrics are computed only on valid-orientation pixels that are also correctly localised as OBs. We additionally report average crispness (AC)~\cite{ye2023delving}, the ratio of post-NMS to pre-NMS edge response mass, indicating whether predictions are intrinsically crisp or rely on NMS thinning. 

\vspace{2pt}
\noindent\textbf{Data and augmentation.}
We use the 390/130 train/test split described in \refsec{obb_sec:dataset_stats}. Each image is annotated at native resolution with a binary OB mask ($B$), an orientation map following the left-hand-rule convention, and a validity mask ($\mathcal{B}_{\text{valid}}$) indicating pixels used for orientation supervision and evaluation. Offline geometric augmentation ($16\times$), together with 248 DIODE hard-negative images containing rich edges but no OBs, yields a training pool of 6{,}488 samples. Test images are evaluated at native resolution.

\vspace{2pt}
\noindent\textbf{Training Protocol.}
Most models are trained under a unified protocol in PyTorch (AdamW, 50K iterations, batch size 8). Orientation loss is applied only to valid-orientation OB pixels, while all annotated OB pixels supervise boundary localisation. MEMO~\cite{Cheng2026MEMO}, DiffusionEdge~\cite{Ye2024DiffusionEdge} and MatchED~\cite{Cetinkaya2026MatchED} keep their original recipes, aligned with ours in iteration budget, split, effective batch size and checkpoint selection; the first two additionally start from task-specific edge pretraining rather than an ImageNet backbone~\cite{deng2009imagenet}, so their results carry an extra intensity-based edge prior.

\begin{table*}[t]
\vspace{-13pt}
\centering
\caption[Main benchmark results on \finalRealBenchmarkName.]{
\textbf{Main benchmark results on \finalRealBenchmarkName{}.}
GFLOPs are measured at $1024 \times 768$ resolution.
For MEMO, the reported GFLOPs denote the \emph{total} inference cost across all 40 UNet forwards.
$^\dagger$ indicates methods re-implemented by us without publicly released code. $^\ddagger$ indicates methods with adapted configuration (see \emph{Supp.}).
Best results within each method group are shown in \textbf{bold}.}
\vspace{-6pt}
\label{tab:realobb_main}
\resizebox{0.8\textwidth}{!}{%
\begin{tabular}{lccc|ccc|c|ccc}
\toprule
\textbf{Model} & \textbf{Venue} & \textbf{Params (M)} & \textbf{GFLOPs} & \textbf{B-ODS$\uparrow$} & \textbf{B-OIS$\uparrow$} & \textbf{B-AP$\uparrow$} & \textbf{AC$\uparrow$} & \textbf{O-ODS$\uparrow$} & \textbf{O-OIS$\uparrow$} & \textbf{O-AP$\uparrow$} \\
\midrule
\multicolumn{11}{l}{\textbf{\textit{Dedicated OB estimation methods}}} \\
DOC-HED~\cite{wang2016doc}             & ECCV'16 & 14.72 & 483.0 & 0.531 & 0.608 & 0.407 & 0.196 & 0.208 & 0.245 & 0.075 \\
DOC-DMLFOV~\cite{wang2016doc}           & ECCV'16 & 20.49 & 2138.8 & 0.362 & 0.392 & 0.277 & 0.205 & 0.135 & 0.147 & 0.045 \\
DOOBNet~\cite{wang2019doobnet}             & ACCV'18 & 32.33 & 295.4 & 0.747 & 0.772 & 0.786 & \textbf{0.741} & 0.396 & 0.416 & 0.264 \\
CCENet$^\dagger$~\cite{lu2019context}     & ICME'19 & 39.32	 & 181.3 & 0.688 & 0.708 & 0.702 & 0.219 & 0.372 & 0.392 & 0.230 \\
OFNet~\cite{lu2019ofnet}                & ICCV'19 & 32.56 & 343.1 & 0.748 & 0.771 & 0.785 & 0.249 & 0.371 & 0.389 & 0.227 \\
P2ORM~\cite{qiu2020P2ORM}                & ECCV'20 & 19.17 & 533.5 & 0.726 & 0.750 & 0.759 & 0.248 & 0.341 & 0.362 & 0.195 \\
OPNet~\cite{feng2021mtorl}                & ICCV'21 & 187.05 & 1773.9 & 0.619 & 0.674 & 0.622 & 0.258 & 0.269 & 0.294 & 0.125 \\
FSINet$^\dagger$~\cite{li2023fsinet}     & VI'23   & 37.44 & 248.9 & 0.745 & 0.768 & \textbf{0.795} & 0.302 & 0.391 & 0.413 & 0.249 \\
OBP-GAN$^\dagger$~\cite{hambarde2024occlusion}    & MM'24   & 165.74 & 924.4 & 0.648 & 0.693 & 0.544 & 0.201 & 0.365 & 0.395 & 0.196 \\
TPENet~\cite{xu2024iobe}$^\ddagger$                & BMVC'25 & 283.23 & 2858.8 & \textbf{0.760} & \textbf{0.776} & 0.691 & 0.318 & \textbf{0.489} & \textbf{0.502} & \textbf{0.303} \\
MoDOT~\cite{xu2026modot}$^\ddagger$ & WACV'26 & 279.50 & 1534.3 &  0.673 &	0.694 &	0.702 &	0.271 &	0.236 &	0.248 &	0.093 \\
\midrule
\multicolumn{11}{l}{\textbf{\textit{Edge detectors with orientation head}}} \\
HED~\cite{Xie2015Holistically}                  & ICCV'15 & 15.03 & 721.7 & 0.601 & 0.665 & 0.499 & 0.222 & 0.305 & 0.336 & 0.145 \\
RCF~\cite{Liu2017Richer}                  & CVPR'17 & 15.12 & 859.7 & 0.475 & 0.568 & 0.319 & 0.187 & 0.212 & 0.252 & 0.069 \\
CASENet~\cite{yu2017casenet}              & CVPR'17 & 24.12 & 365.9 & 0.618 & 0.668 & 0.562 & 0.209 & 0.316 & 0.349 & 0.165 \\
BDCN~\cite{he2019bdcn}                 & CVPR'19 & 16.61 & 1106.1 & 0.578 & 0.647 & 0.503 & 0.244 & 0.256 & 0.291 & 0.111 \\
DexiNed~\cite{poma2020biped}              & WACV'20 & 21.24 & 753.3 & 0.450 & 0.529 & 0.343 & 0.226 & 0.192 & 0.222 & 0.069 \\
PiDiNet~\cite{Su2021Pixel}              & ICCV'21 & 0.94 & 362.6 & 0.553 & 0.614 & 0.466 & 0.242 & 0.234 & 0.261 & 0.096 \\
CATS~\cite{huan2021unmixing}                 & TPAMI'21   & 15.16 & 927.8 & 0.733 & 0.766 & 0.759 & \textbf{0.599} & 0.355 & 0.379 & 0.194 \\
LDC-B5~\cite{soria2022ldc}               & Access'22 & 0.84 & 103.1 & 0.534 & 0.582 & 0.461 & 0.181 & 0.228 & 0.249 & 0.097 \\
EDTER~\cite{Pu2022EDTER}                & CVPR'22 & 357.69 & 5403.5 & 0.324 & 0.361 & 0.231 & 0.207 & 0.149 & 0.159 & 0.054 \\
EDTER (stage~2)     & CVPR'22 & 533.76 & 13462.6 & 0.443 & 0.515 & 0.358 & 0.201 & 0.191 & 0.219 & 0.074 \\
RankED~\cite{Cetinkaya2024RankED}               & CVPR'24 & 120.48 & 468.3 & 0.667 & 0.694 & 0.647 & 0.200 & 0.449 & 0.472 & 0.311 \\
DDN~\cite{Li2025Doubly}                  & NC'25   & 56.26 & 551.9 & \textbf{0.764} & \textbf{0.787} & \textbf{0.818} & 0.498 & \textbf{0.496} & \textbf{0.514} & \textbf{0.390} \\
NBED~\cite{li2026nbed}                 & SPIC'26 & 66.57 & 967.4 & 0.630 & 0.688 & 0.462 & 0.254 & 0.391 & 0.445 & 0.194 \\
\midrule
\multicolumn{11}{l}{\textbf{\textit{Edge detectors, boundary only}}} \\
HED~\cite{Xie2015Holistically}                  & ICCV'15 & 14.72 & 482.9 & 0.665 & 0.711 & 0.617 & 0.259 & -- & -- & -- \\
RCF~\cite{Liu2017Richer}                  & CVPR'17 & 14.80 & 619.8 & 0.644 & 0.697 & 0.594 & 0.229 & -- & -- & -- \\
CASENet~\cite{yu2017casenet}              & CVPR'17 & 23.51 & 129.5 & 0.674 & 0.717 & 0.675 & 0.243 & -- & -- & -- \\
BDCN~\cite{he2019bdcn}                 & CVPR'19 & 16.30 & 866.1 & 0.637 & 0.696 & 0.601 & 0.253 & -- & -- & -- \\
DexiNed~\cite{poma2020biped}              & WACV'20 & 20.93 & 514.5 & 0.587 & 0.646 & 0.506 & 0.271 & -- & -- & -- \\
PiDiNet~\cite{Su2021Pixel}              & ICCV'21 & 0.71 & 119.1 & 0.625 & 0.684 & 0.606 & 0.244 & -- & -- & -- \\
CATS~\cite{huan2021unmixing}                 & TPAMI'21  & 14.85 & 687.8 & 0.746 & 0.777 & 0.772 & 0.569 & -- & -- & -- \\
LDC-B5~\cite{soria2022ldc}               & Access'22 & 0.79 & 43.7 & 0.626 & 0.688 & 0.595 & 0.220 & -- & -- & -- \\
EDTER~\cite{Pu2022EDTER}                & CVPR'22 & 357.45 & 5169.3 & 0.592 & 0.670 & 0.548 & 0.206 & -- & -- & -- \\
EDTER (stage~2)     & CVPR'22 & 533.69 & 13345.9 & 0.649 & 0.708 & 0.581 & 0.259 & -- & -- & -- \\
DiffusionEdge~\cite{Ye2024DiffusionEdge}        & AAAI'24  & 297.70 & 14153.3 & 0.770 & 0.782 & 0.695 & 0.874 & -- & -- & --  \\
RankED~\cite{Cetinkaya2024RankED}               & CVPR'24 & 120.11 & 234.5 & 0.670 & 0.696 & 0.607 & 0.212 & -- & -- & -- \\
DDN~\cite{Li2025Doubly}                  & NC'25   & 56.15 & 492.6 & \textbf{0.812} & \textbf{0.823} & \textbf{0.841} & 0.596 & -- & -- & -- \\
NBED~\cite{li2026nbed}                 & SPIC'26 & 66.50 & 905.0 & 0.745 & 0.783 & 0.722 & 0.333 & -- & -- & -- \\
MEMO~\cite{Cheng2026MEMO}                 & CVPR'26 & 241.32 & 600752.8 & 0.705 & 0.727 & 0.334 & \textbf{0.888} & -- & -- & --  \\
MatchED~\cite{Cetinkaya2026MatchED}              & CVPR'26 & 0.73 & 80.71 & 0.736 &	0.770	& 0.761 & 	0.835 & -- & -- & --  \\
\bottomrule
\end{tabular}%
}
\vspace{-13pt}
\end{table*}

\subsection{OB and Edge Benchmark Results}
\label{obb_sec:results}

Table~\ref{tab:realobb_main} reports the main benchmark results:

\noindent\textbf{Dedicated OB estimation methods.}
OB methods have advanced substantially since early CNN detectors: DOC-HED and DOC-DMLFOV reach only $0.531$/$0.362$ B-ODS, whereas DOOBNet, OFNet, FSINet, and TPENet now cluster around $0.745$--$0.760$. Within this group, TPENet leads on both boundary and orientation metrics, while FSINet is best on boundary AP and DOOBNet on crispness. 

\noindent\textbf{Edge detectors as OB baselines.}
Modern edge detectors remain strong OB baselines. With a lightweight orientation head, DDN attains the best scores of \emph{any} method on both boundary and orientation metrics, though its margin over the best dedicated method (TPENet) is now narrow (B-ODS $0.764$ vs.\ $0.760$). RankED also shows strong orientation despite more modest localisation. In the boundary-only setting, DDN improves further (B-ODS $0.812$). 
Attaching the same untuned head lowers B-ODS for every edge detector, from negligible (RankED) to severe (EDTER), and crispness for 11 of 13. These findings highlight the preservation of pure edge-model accuracy under joint orientation supervision as an important direction for future research.

\noindent\textbf{Orientation gap and crispness.}
Two patterns hold across all groups. First, orientation scores trail boundary scores for every method (\eg, DDN O-ODS $0.496$ vs.\ B-ODS $0.764$): accurate cross-boundary depth-ordering reasoning is far from solved even when boundaries are well localised. Second, crispness is largely decoupled from PR metrics---MEMO and DiffusionEdge produce the sharpest edges yet do not lead on PR metrics (MEMO's B-AP is only $0.334$), whereas DDN and CATS combine moderate-to-high crispness (AC $\approx0.50$--$0.60$) with strong PR. 
Overall, the strongest oriented-OB performance on \finalRealBenchmarkName{} comes from a repurposed edge detector (DDN), closely followed by TPENet, and the dominant remaining bottleneck is orientation rather than boundary localisation.

\subsection{Probing Monocular Depth for OB Fidelity}
\label{sec:depth-edge-protocols}

\begin{table*}[t]
\vspace{-13pt}
\centering
\caption[Boundary localisation on the GT-depth subset of \finalRealBenchmarkName.]{%
  \textbf{Boundary localisation of monocular depth predictions on the GT-depth subset of \finalRealBenchmarkName{} ($N{=}209$).} F1$_w$ and R$_w$ are threshold-weighted over the 10 ratio thresholds of~\cite{bochkovskii2025depthpro} (see \emph{Supp.}). Best per column in \textbf{bold}.}
  \vspace{-6pt}
\label{tab:depth_edge_gtsubset}
\resizebox{0.8\textwidth}{!}{%
\begin{tabular}{lcc|cc|cc|ccc}
  \toprule
  & \multicolumn{2}{c|}{\textbf{Canny strict-pixel~\cite{xu2026modot}}}
  & \multicolumn{2}{c|}{\textbf{Canny + Chamfer~\cite{qiu2020P2ORM}}}
  & \multicolumn{2}{c|}{\textbf{SI-BF1~\cite{bochkovskii2025depthpro}}}
  & \multicolumn{3}{c}{\textbf{SEval on $\mathcal{B}_{\text{valid}}$}} \\
  \cmidrule(lr){2-3} \cmidrule(lr){4-5} \cmidrule(lr){6-7} \cmidrule(lr){8-10}
  \textbf{Model}
    & \textbf{R$\uparrow$} & \textbf{F1$\uparrow$}
    & \textbf{dbe$_{\text{acc}}\downarrow$} & \textbf{dbe$_{\text{com}}\downarrow$}
    & \textbf{F1$_{w}\uparrow$} & \textbf{R$_{w}\uparrow$}
    & \textbf{B-ODS$\uparrow$} & \textbf{B-OIS$\uparrow$} & \textbf{B-AP$\uparrow$} \\
  \midrule
  DA v1~\cite{yang2024depthanything}
    & 0.104 & 0.132 & 2.48 & 6.13 & 0.119 & 0.092 & 0.421 & 0.237 & 0.156 \\
  DA v2~\cite{yang2024depthanythingv2}
    & 0.137 & 0.168 & 2.10 & 5.49 & \textbf{0.174} & 0.147 & 0.462 & 0.314 & 0.145 \\
  DepthPro~\cite{bochkovskii2025depthpro}
    & 0.204 & 0.252 & 1.89 & 5.10 & 0.105 & 0.077 & \textbf{0.549} & \textbf{0.553} & \textbf{0.334} \\
  MoGe-2~\cite{wang2025mogev2}
    & 0.182 & 0.227 & 2.35 & 5.36 & 0.097 & 0.071 & 0.538 & 0.494 & 0.311 \\
  PPD~\cite{xu2025ppd}
    & \textbf{0.220} & \textbf{0.260} & \textbf{1.85} & \textbf{4.94} & 0.109 & \textbf{0.168} & 0.264 & 0.179 & 0.053 \\
  InfiniDepth~\cite{yu2026infinidepth}
    & 0.201 & 0.241 & 2.22 & 5.35 & 0.161 & 0.128 & 0.523 & 0.532 & 0.236 \\
  \bottomrule
\end{tabular}}
\vspace{-13pt}
\end{table*}

We evaluate six state-of-the-art monocular depth estimators with released weights: Depth-Anything (DA) v1/v2 (Large)~\cite{yang2024depthanything,yang2024depthanythingv2}, MoGe-2~\cite{wang2025mogev2}, DepthPro~\cite{bochkovskii2025depthpro}, PPD~\cite{xu2025ppd}, and InfiniDepth~\cite{yu2026infinidepth}. Predictions are bilinearly upsampled to input resolution when needed.
Since a subset of orientation labels is generated with the aid of predicted depth from MoGe-2 and PPD (\refsec{obb_sec:gen_ori}), we report the main depth-based evaluation only on the GT-depth portion of \finalRealBenchmarkName{} ($N=209$) to avoid potential information leakage.
Unlike OB and edge detectors, which are evaluated on the full OB annotation set $\mathcal{B}$, depth-based orientation evaluation is computed on the valid-orientation subset $\mathcal{B}_{\text{valid}}$, where each evaluated OB pixel has a measurable cross-boundary depth difference. 
On this portion, $\mathcal{B}_{\text{valid}}$ covers 46.2\% of OB pixels and is the most favourable target for depth methods.

We use \textbf{\emph{five}} complementary protocols (full definitions in \emph{Supp.}):
\textbf{(i)} strict-pixel Canny matching on normalised depth maps~\cite{xu2026modot};
\textbf{(ii)} Canny-based Chamfer distances~\cite{qiu2020P2ORM};
\textbf{(iii)} Scale-Invariant Boundary F1 (SI-BF1)~\cite{bochkovskii2025depthpro};
\textbf{(iv)} SEval (Section~\ref{obb_sec:experimentalSetup}) on the extracted depth-ratio edge maps in (iii), together with orientations generated from each model's depth, evaluating orientation jointly with boundary localisation;
and \textbf{(v)} a cross-boundary depth-ordering analysis that feeds the \emph{GT} OBs and each model's predicted depth into our orientation-generation algorithm, evaluating orientation in isolation from boundary localisation. For (v), \emph{coverage}, \emph{d-ord (cond.)}, and \emph{d-ord (joint)} report the committed fraction of $\mathcal{B}_{\text{valid}}$, the correct ordering ($<\!90^{\circ}$ error) among committed pixels, and the correct ordering over all of $\mathcal{B}_{\text{valid}}$ (non-commitment counts as wrong), respectively.

\begin{table}[ht]
\centering
\caption[Cross-boundary depth ordering recovered from predicted depth on the GT-depth subset.]{
  \textbf{Cross-boundary depth-ordering fidelity of predicted depth on the GT-depth subset of \finalRealBenchmarkName{}.}}
  \vspace{-6pt}
\label{tab:orientation_pathb}
\resizebox{\columnwidth}{!}{
\begin{tabular}{lccc|ccc}
  \toprule
  & \multicolumn{3}{c|}{\textbf{Orientation SEval}}
  & \multicolumn{3}{c}{\textbf{Cross-boundary depth ordering}} \\
  \cmidrule(lr){2-4} \cmidrule(lr){5-7}
  \textbf{Model}
    & \textbf{O-ODS$\uparrow$} & \textbf{O-OIS$\uparrow$} & \textbf{O-AP$\uparrow$}
    & \textbf{coverage$\uparrow$} & \textbf{d-ord (cond.)$\uparrow$} & \textbf{d-ord (joint)$\uparrow$} \\
  \midrule
  DA v1 & 0.218 & 0.125 & 0.039 & 0.744 & 0.893 & 0.666 \\
  DA v2 & 0.245 & 0.166 & 0.037 & 0.802 & 0.895 & 0.722 \\
  DepthPro & \textbf{0.312} & \textbf{0.330} & \textbf{0.104} & \textbf{0.828} & 0.892 & \textbf{0.744} \\
  MoGe-2 & 0.306 & 0.292 & 0.097 & 0.813 & 0.889 & 0.732 \\
  PPD & 0.096 & 0.066 & 0.008 & 0.809 & 0.889 & 0.725 \\
  InfiniDepth & 0.296 & 0.314 & 0.072 & 0.791 & \textbf{0.899} & 0.710 \\
  \bottomrule
\end{tabular}}
\vspace{-13pt}
\end{table}

Tables~\ref{tab:depth_edge_gtsubset} and~\ref{tab:orientation_pathb} show that, even on $\mathcal{B}_{\text{valid}}$---the subset deliberately chosen to favour depth---visual sharpness is not a proxy for OB fidelity. Depth-derived edges stay far from pixel-accurate: the best strict-pixel Canny F1 is only $0.260$ (PPD), Chamfer errors are $1.85$--$6.13$ pixels, and no estimator dominates across protocols. 
Orientation is weaker still: the best orientation SEval reaches only O-ODS $0.312$ (DepthPro), and once localisation error is removed, two-thirds of the residual error comes from pixels where predicted depth shows no measurable gap at all rather than the wrong side (joint $74.4\%$, conditional $89.2\%$).
State-of-the-art monocular depth estimators do not recover definition-consistent OBs---a result that establishes \finalRealBenchmarkName{} as a real-world depth-boundary testbed and motivates the open problems that we discuss in the next section.

\subsection{Discussion, Insights and Limitations}
\label{obb_sec:discussion}

Taken together, the experiments answer the three questions of \refsec{obb_sec:experiments}. Under a single unified protocol, modern edge detectors can already serve as strong OB estimators, while leaving substantial room for further OB-specific adaptation, so \emph{boundary localisation} is within reach of modern dense predictors. Beyond localisation, the problem stays challenging: occlusion \emph{orientation} trails boundary quality for every method, and even state-of-the-art monocular depth estimators fail to deliver pixel-accurate OBs or measurable cross-boundary geometry on the depth-favourable subset $\mathcal{B}_{\text{valid}}$. That \finalRealBenchmarkName{} cleanly separates these regimes is what such a benchmark should do.

That edge detectors match or surpass dedicated OB methods reflects a difference in capacity allocation rather than intrinsic superiority: edge networks learn strong boundary priors from dense supervision that transfer to surface-visibility discontinuities, whereas many OB-specific designs were developed on category-dependent data (\eg, PIOD) without self-occlusion.
Since orientation can be attached to an edge detector at an architecture-dependent cost (B-ODS $-0.013$ for CATS but $-0.268$ for EDTER), a competitive oriented-OB system does not need to begin from an OB-specific design.

The orientation gap is the dominant open problem, and two factors sustain it.
First, the OB evidence is often simply absent: the best estimator's depth resolves no cross-boundary gap at $17\%$ of the reliably measurable pixels while ordering the rest correctly $89\%$ of the time---the bottleneck is missing evidence at true OBs, not wrong ordering.
Second, orientation supervision is harder to obtain than boundary supervision---for sources without GT depth we derive it from predicted depth and manual verification, leaving residual label noise on subtle, low-contrast boundaries. Our validity-aware design mitigates but does not remove this noise, so separating model error from label error remains an open question that \finalRealBenchmarkName{} is built to study.

\vspace{3pt}
\noindent\textbf{Limitations.} \finalRealBenchmarkName{} matches widely used manually annotated benchmarks such as BSDS500~\cite{arbelaez2010contour} in scale at over four times its pixel count, and additionally serves as a testbed for depth methods. But it is not designed as a large pretraining corpus: annotation cost bounds its scale and excludes outdoor vegetation and dense clutter, whose occlusion structure cannot be delineated without undermining definition consistency. Expanding both is future work.

\section{Conclusion}
\label{obb_sec:conclusion}

We introduced \finalRealBenchmarkName{}, a definition-consistent benchmark for real-world oriented
occlusion boundary estimation, providing carefully verified binary OB annotations, validity-aware orientation labels, and a unified evaluation protocol.
Our experiments show that modern edge detectors are strong OB localisation baselines, surpassing dedicated OB architectures on both boundary and orientation metrics, yet accurate occlusion orientation reasoning remains a bottleneck across all methods.
Probing monocular depth estimators further reveals that visually sharp depth maps do not translate to pixel-accurate OBs and often leave no measurable depth step at true OBs.
By releasing the dataset, annotation guidelines, orientation-generation algorithm, and evaluation tools, we hope \finalRealBenchmarkName{} will serve as a rigorous reference benchmark for oriented OB estimation and a complementary testbed for boundary fidelity in monocular depth and broader geometry-aware low-level vision.

\clearpage

{
    \small
    \bibliographystyle{ieeenat_fullname}
    \bibliography{egbib}
}

\end{document}